\documentclass[letterpaper, 10 pt, conference]{ieeeconf}  % Comment this line out if you need a4paper

\IEEEoverridecommandlockouts                              % This command is only needed if
\makeatletter
\let\NAT@parse\undefined
\makeatother
\usepackage[numbers,sort&compress]{natbib}
\usepackage{fancyhdr}
\fancypagestyle{withfooter}{
  
  \fancyfoot[C]{\footnotesize Accepted to the IEEE ICRA Workshop on Field Robotics 2024}
}
\usepackage[pdftex]{graphicx}
\usepackage{amsmath}
\usepackage{amssymb}  % assumes amsmath package installed
\usepackage{subfigure}
\usepackage{multirow}
\usepackage{array,booktabs}
\usepackage{diagbox}
\usepackage{balance}
\usepackage{xspace}
\usepackage{algorithm}
\usepackage{algpseudocode}
\usepackage{adjustbox}
\usepackage{romannum}
 \usepackage[final]{hyperref}
 \hypersetup{
     colorlinks=true,
     linkcolor=blue,
     filecolor=magenta,
     urlcolor=blue,
     citecolor=black
 }
\usepackage[labelfont=bf, font=small]{caption}
\usepackage{./rpm_packages/rpm_SIunits}
\usepackage{./rpm_packages/rpm_acronyms}
\usepackage{./rpm_packages/rpm_math}
\usepackage{./rpm_packages/rpm_misc}

\usepackage{soul,color}
\usepackage{lipsum}

\usepackage{xcolor}

\usepackage{tikz} % \checkmark
\usepackage{subcaption}
\usepackage{caption}
\title{\LARGE \bf Toward Robust LiDAR based 3D Object Detection via \\Density-Aware Adaptive Thresholding}     

\author{Eunho Lee${}^{1}$, Minwoo Jung${}^{2}$ and Ayoung Kim${}^{2*}$
\thanks{$^\dagger$This work was supported by Institute of Information \& communications Technology Planning \& Evaluation (IITP) grant funded by the Korea government(MSIT) No.2022-0-00480, Development of Training and Inference Methods for Goal-Oriented Artificial Intelligence Agents}
\thanks{$^{1}$E. Lee is with the Interdisciplinary Program in Artifical Intelligence, SNU, Seoul, S. Korea {\tt\small eunho1124@snu.ac.kr}}%
\thanks{$^{2}$M. Jung and A. Kim are with the Dept. of Mechanical Engineering, SNU, Seoul, S. Korea {\tt\small [moonshot, ayoungk]@snu.ac.kr}}%
}
\begin{document}

%\onecolumn
\maketitle
\thispagestyle{withfooter}
\pagestyle{withfooter}

\begin{abstract}

Robust 3D object detection is a core challenge for autonomous mobile systems in field robotics. To tackle this issue, many researchers have demonstrated improvements in 3D object detection performance in datasets. However, real-world urban scenarios with unstructured and dynamic situations can still lead to numerous false positives, posing a challenge for robust 3D object detection models. This paper presents a post-processing algorithm that dynamically adjusts object detection thresholds based on the distance from the ego-vehicle. 3D object detection models usually perform well in detecting nearby objects but may exhibit suboptimal performance for distant ones. While conventional perception algorithms typically employ a single threshold in post-processing, the proposed algorithm addresses this issue by employing adaptive thresholds based on the distance from the ego-vehicle, minimizing false negatives and reducing false positives in urban scenarios. The results show performance enhancements in 3D object detection models across a range of scenarios, not only in dynamic urban road conditions but also in scenarios involving adverse weather conditions.

\end{abstract}
\section{Introduction}
\label{sec:intro}

%FIGURE

3D object detection is one of the fundamental components of autonomous mobile systems in field robotics. This system often handles the unstructured and dynamic real-world environments that invoke unpredictable situations \cite{wijayathunga2023challenges, ginerica2024vision, itkina2019dynamic}. However, 3D object detectors trained in datasets perform less in various real-world environments than in datasets, leading to system malfunction. 

Robust 3D object detection is crucial to function for their purpose in highly dynamic environments like urban roads, where dynamic objects, unpredictable obstacles, and sensor noise present significant challenges. While urban roads might seem structured, they are also one of the key environments that field robotics must handle because they are dynamic and challenging \cite{9046805}. These environments are filled with moving vehicles, pedestrians, and ghost obstacles worsened by adverse weather \cite{9564505, Hahner_2021_ICCV}. This leads to numerous false positives, potentially causing sudden stops and fatal accidents. Therefore, developing a robust perception module that minimizes false positives on urban roads is a key research focus in field robotics.

% Robust 3D object detection in unstructured and dynamic environments is essential to function for their purpose. 
% These environments often contain numerous dynamic objects and unpredictable obstacles, challenging robustness. Urban roads serve as a suitable example of such dynamic situations. While urban roads might seem structured, they are also one of the key environments that field robotics must handle because they are dynamic and challenging \cite{9046805}. There are many moving objects like vehicles and pedestrians, unpredictable obstacles such as bushes and road signs, and ghost obstacles due to sensor noise. Moreover, adverse weather conditions like fog and rain in real-world urban roads \cite{9564505, Hahner_2021_ICCV} induce sensor noise and unexpected detections. These scenarios often lead to numerous false positives, which can cause critical issues, including sudden stops and traffic accidents. Hence, developing a robust perception module that minimizes false positives in urban roads is one of the important research within field robotics.

\begin{figure}[!t]
    \centering
    \includegraphics[width=1\columnwidth]{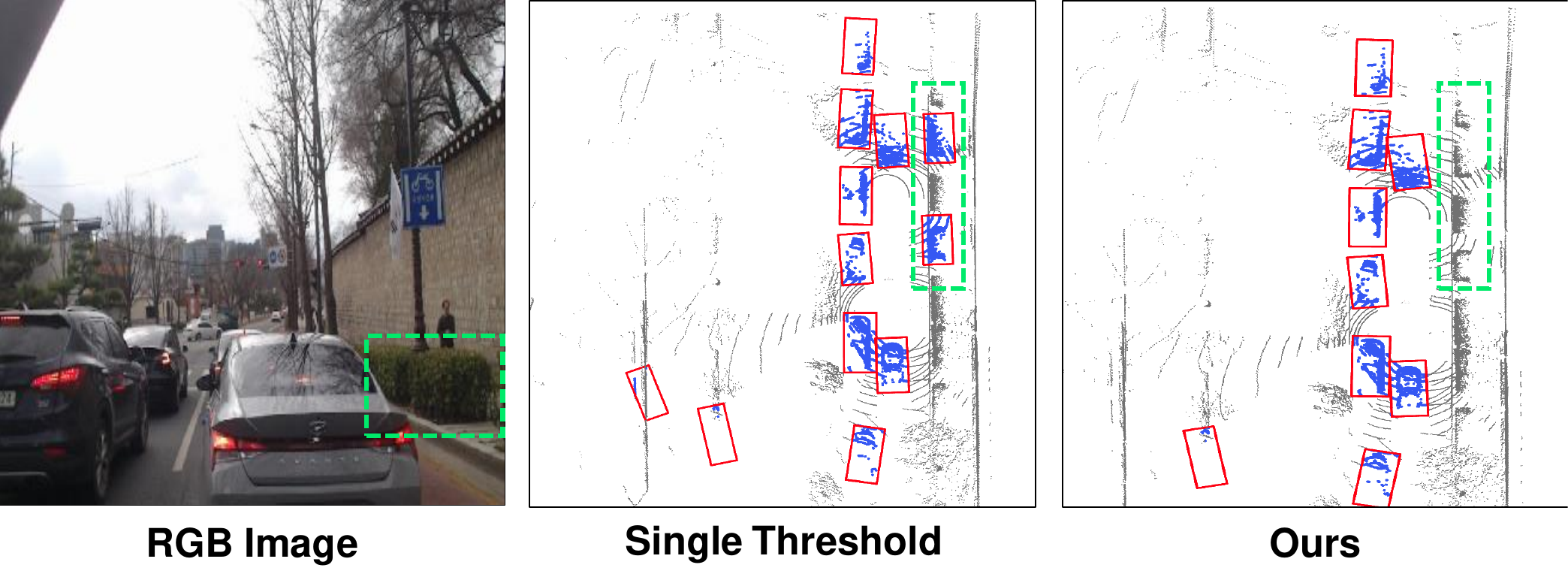}
    \caption{
     The dynamic environment involves moving vehicles, pedestrians, and obstacles such as bushes and road signs (RGB image). 3D object detection using a single threshold misclassifies the bushes as vehicles, which results in false positives (single threshold). Our method significantly improves the robustness of 3D object detection in complex urban scenarios by effectively minimizing false detections, thus enhancing autonomous driving systems' overall performance and safety (ours).  
    }
    \label{fig: Qualitative Result}
    \vspace{-3mm}
\end{figure}

Traditionally, 3D object detection usually utilizes LiDAR, offering high precision and efficient 3D environmental data for robust perception.
LiDAR-based 3D object detection models \cite{zhou2018voxelnet, ren2015faster, yan2018second,   lang2019pointpillars, shi2019pointrcnn, shi2020pv, duan2019centernet} use 3D point clouds to predict an object’s class, location, and confidence scores, which are then refined by confidence score-based post-processing using a single threshold hyper-parameter. Detection accuracy varies with object distance due to sensor characteristics like resolution and range and training dataset diversity. Objects near the ego-vehicle have higher accuracy and confidence due to the increased density of point clouds, whereas distant objects have lower recall and confidence scores due to the corresponding decrease in point cloud density. These observations indicate that for autonomous mobile systems to drive safely on real roads, it is more suitable to prioritize precision for objects closer and recall for objects farther away. Consequently, employing a single threshold in post-processing is inadequate for autonomous mobile systems operating across diverse real-world environments.

% LiDAR sensors have been widely utilized for autonomous mobile systems due to their high precision and efficiency in providing detailed 3D environmental data, enabling robust perception in unstructured and dynamic environments. 
% LiDAR-based 3D object detection models [4–10] take 3D point clouds and predict an object’s class, location, and confidence scores. Then, these predictions are processed through confidence score-based post-processing to produce the final results, where a single threshold value is often employed as a hyper-parameter for handling confidence scores. Notably, due to the inherent characteristics of sensors, including their resolution and range, coupled with the diversity of 3D object detection datasets used for training, a noticeable variance exists in detection accuracy upon the distance of the detected objects. Specifically, objects closer to the ego-vehicle are generally detected with higher accuracy, characterized by high recall and confidence scores, due to the increased density of point clouds. In contrast, more distant objects tend to show reduced recall and confidence scores due to the corresponding decrease in point cloud density.
% These observations indicate that for autonomous vehicles to drive safely on real roads, it is more suitable to prioritize precision for objects closer and recall for objects farther away. 

    %FIGURE
\begin{figure*}[t!]
    \centering
    \includegraphics[width=1\textwidth]
    {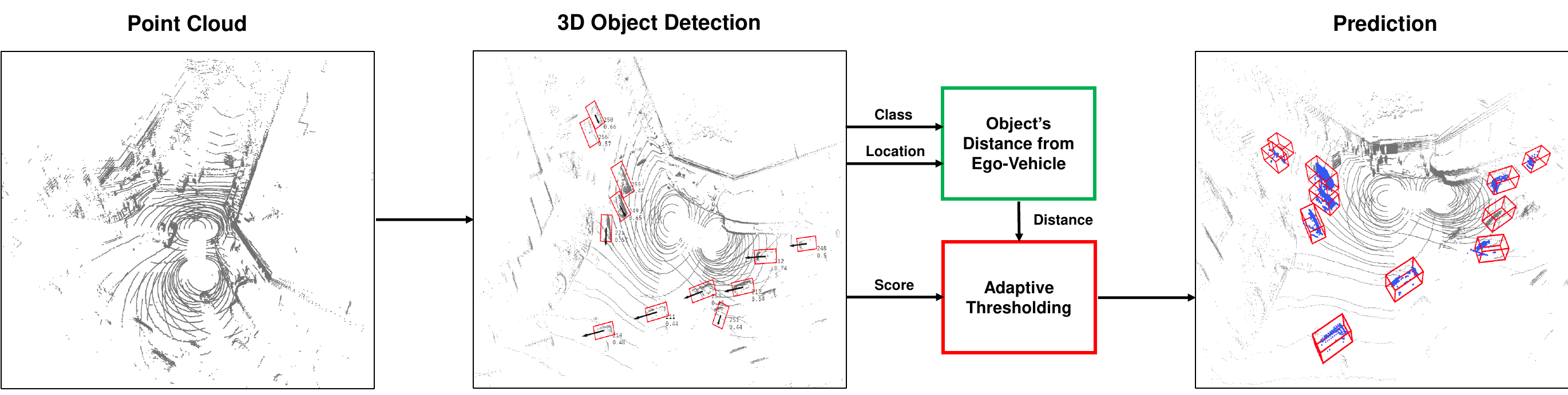}
    \caption{
    Framework of the proposed algorithm. 3D Detectors \cite{zhou2018voxelnet, ren2015faster, yan2018second, lang2019pointpillars, shi2019pointrcnn, shi2020pv, duan2019centernet} normally apply the post processing using the single threshold. In contrast, the adaptive thresholding. 
    }
    \label{fig: Overview}
    \vspace{-0.3cm}
\end{figure*}

While existing methods on the post-processing algorithm utilized adaptive thresholding \cite{zhang2021moving, lin2018moving} have explored adaptive thresholding techniques, their focus has remained on 2D image processing, which does not fully satisfy 3D object detections. To tackle this issue, our work introduces a novel adaptive thresholding algorithm specifically designed for 3D object detection, making it better suited for implementation in autonomous mobile systems. This innovation also offers a straightforward integration process that does not require extra training or add unnecessary complexity to existing detection frameworks. By optimizing the balance between minimizing false positives for nearby objects and lowering missed detections for distant objects, our approach enhances the overall efficacy of 3D object detection in datasets and real-world scenarios. This contributes to safer and more reliable autonomous navigation, as demonstrated in our qualitative analysis presented in \figref{fig: Qualitative Result}.

Differing from previous methods, our method presents the following contributions: 

\begin{itemize}
    \item {By applying adaptive thresholding based on the distance from the ego-vehicle in the post-processing of 3D object detection models, we significantly reduce false positives and false negatives according to distance. This improvement strengthens the robustness of models and makes the model more suitable for autonomous mobile systems by enabling more stable driving.
    }
    
    \item {
    The proposed algorithm simplifies the integration into various 3D object detection frameworks and effectively boosts detection performance in terms of \ac{mAP} and the balance between Recall and Precision. Our experiments across multiple frameworks have confirmed these improvements.
    }
    
    \item {The reduction in misidentification issues within the model has been qualitatively validated based on actual driving data from diverse urban roads and under varying weather conditions.}

\end{itemize}

\section{related work}
\label{sec:relatedwork}
\subsection{Adaptive Thresholding in Object Detection}

Previous researches on adaptive thresholding \cite{zhang2021moving, lin2018moving} have been conducted for detecting moving objects in 2D images. \citeauthor{zhang2021moving} \cite{zhang2021moving} enhanced detection accuracy by employing adaptive thresholding based on distance. This was achieved by detecting objects within an image and subsequently estimating the objects' states and distances through pixel comparison with previous frames. \citeauthor{lin2018moving} \cite{lin2018moving} utilizes adaptive thresholding to determine thresholds for distinguishing between moving and stationary objects during depth estimation, a critical step in converting images into a 3D representation. This approach involves considering features within the image in pairs, with the threshold adjusting dynamically based on the distance between these features, thereby facilitating the determination of an object's motion status. 
These studies have focused on 2D images, making them unsuitable for 3D object detection models required in autonomous driving, which necessitates representation within a 3D space for various conditions and object recognition.

\subsection{LiDAR based 3D Object Detection}
3D object detection can be categorized into LiDAR-based, Camera-based, Radar-based, and Sensor-Fusion-based modalities. This paper is dedicated solely to exploring LiDAR-based 3D object detection strategies. Within the spectrum of LiDAR-based methodologies, we classify approaches into voxel, pillar, and raw point cloud-based. 

%%수정필요
VoxelNet \cite{zhou2018voxelnet} aggregates and downsamples 3D point clouds into voxels for feature representation and executes object detection via 3D convolution operations facilitated by a Region Proposal Network (RPN) \cite{ren2015faster}. SECOND \cite{yan2018second} introduces an effective voxel computation strategy for handling the sparse attributes present in 3D environments. PointPillars \cite{lang2019pointpillars} adopts a pillar representation to facilitate 2D convolution operations, optimizing computational and memory efficiencies while ensuring robust performance. PointRCNN \cite{shi2019pointrcnn} distinguishes objects from the background using raw point clouds, employing this distinction for object detection to compromise high precision. PV-RCNN \cite{shi2020pv} leverages both raw point clouds and voxels for object detection, preserving the intrinsic properties of point clouds despite significant computational requirements, thus demonstrating remarkable efficiency. These methods rely on anchor-based detection paradigms. Progressing beyond, \citeauthor{duan2019centernet} \cite{duan2019centernet} proposed CenterNet, advocating an anchor-free paradigm, thereby challenging traditional anchor-dependent frameworks. 

Most 3D object detection models are trained and evaluated on specific datasets representing typical, non-adversarial road environments. However, these models struggle to adequately address the more diverse and challenging situations autonomous vehicles encounter on actual urban roads. On actual roads, varying weather conditions such as rain, snow, and fog can introduce noise into sensor data, and elements like bushes or road signs may be mistakenly identified as vehicles or obstacles, leading to false positives. We propose a method that enables autonomous vehicles to adaptively handle false positives, ensuring stable driving on real urban roads.

\section{Adaptive Thresholding Framework}
\label{sec:method}

We illustrate our framework in \figref{fig: Overview}. The 3D object detection model that receives the point clouds as input detects surrounding objects and outputs the class, location, and confidence score. Then, the adaptive thresholding module receives the distance from the ego-vehicle for each class of objects and the confidence score as input and finally predicts the 3D bounding boxes of objects accurately. 
This study used PointPillars \cite{lang2019pointpillars} as our 3D object detection model. We analyze how distinct single threshold values, such as 0.3, 0.5, and 0.7, impact the distance-based performance results of the 3D detector within the Kitti 3D object detection dataset \cite{geiger2012we}. 
In analyzing detection performance with a threshold of 0.5, it is observed that the 3D object detection algorithm effectively identifies objects closer to the ego-vehicle (d $<$ 30m), resulting in the number of detections that surpass the ground truth counts. However, its performance drops over extended ranges (d $>$ 30m), where fewer objects are detected than the ground truth. It highlights a variable recall rate with higher recall at closer proximities and lower recall at extended distances. The inverse relationship between Recall and Precision indicates lower precision at close distances and higher precision at longer distances. 
To enhance precision for objects at closer distances, a higher threshold of 0.7 was applied, leading to a general decrease in detection counts. Notably, the detection rate for objects beyond 40m dropped significantly, resulting in lower recall and increased false negatives. Conversely, a lower threshold of 0.3 was employed to increase recall for distant objects, which improved recall and increased the number of detected objects within the 10 to 40m range beyond the ground truth, thereby reducing overall precision. 
\begin{figure}[t!]
    \centering
    \includegraphics[width=0.85\columnwidth]{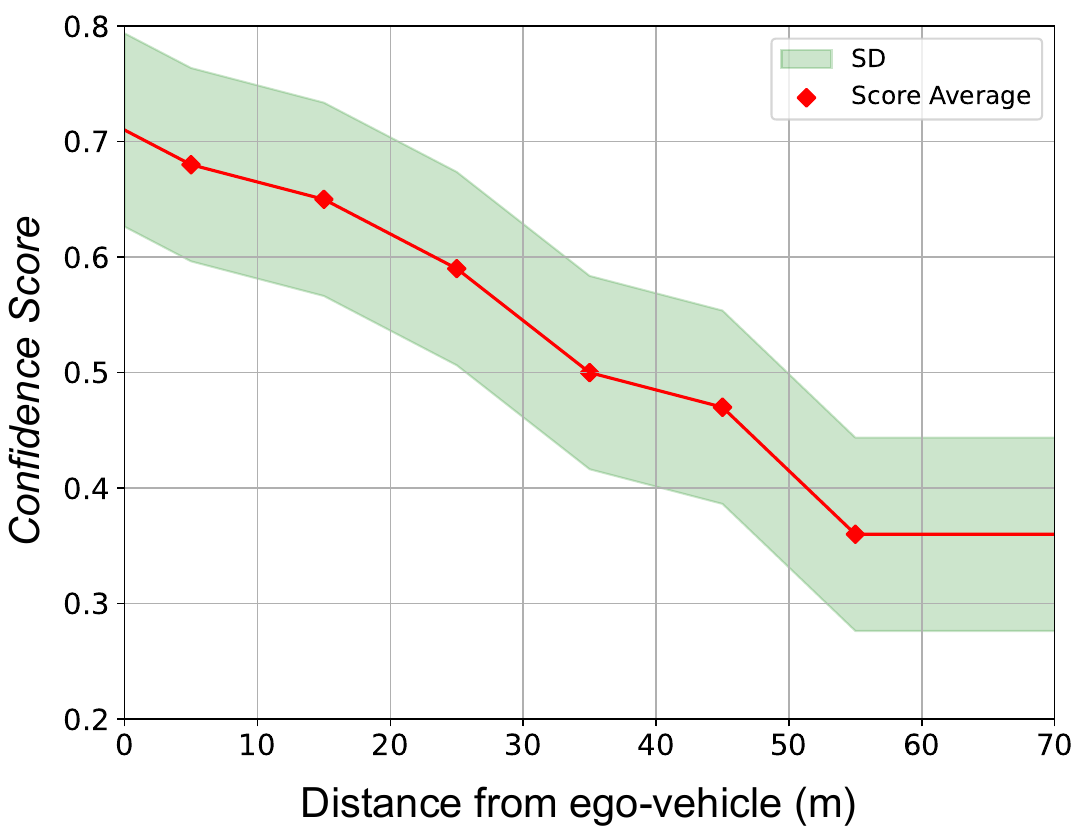}
    \caption{
Confidence score tendency with a single threshold (=0.5). The 
red dots represent the score mean values at each distance (10 
m), and the green-shaded area indicates the standard deviation.}
\label{fig: Graph}
\vspace{-0.5cm}
\end{figure}

The analytical outcomes, employing single thresholds of 0.5 for proximal distances (below 40m) and 0.3 for more extended distances (above 40m), facilitated the computation of the standard deviation across intervals of 10m, as described in \figref{fig: Overview}. After an examination of confidence score trends depicted in \figref{fig: Overview}, it was inferred that a quadratic function most appropriately captures the variability of confidence scores across distances, leading to the formulation of the distance-based adaptive thresholding equation as described in \equref{eq: one}. 

\begin{equation}
\label{eq: one}
\text { Score Threshold }=\left\{\begin{array}{c}
\alpha d^2+\beta d+\gamma(0<d \leq \delta) \\
k(d>\delta)
\end{array}\right.
\end{equation}

This equation, \(d\), represents the distance from the ego-vehicle to the object, measured in meters. The parameters \( \alpha \), \( \beta \), and \( \gamma \) determine the shape of the quadratic curve that models the variation of confidence scores with distance. \( \delta \) is the parameter that defines the maximum distance at which the algorithm is applicable, while \( k \) represents the constant value of the quadratic curve at a distance \( \delta \). The value of \( \delta \) may vary depending on the detection range capabilities of the LiDAR and the performance of the 3D object detection models. To prevent the confidence threshold from decreasing too much for distances beyond \( \delta \), thus increasing the risk of false positives, \( k \) is set accordingly. These parameters enable the application of the algorithm not only to the PointPillars \cite{lang2019pointpillars} model but also across various 3D object detection models.

\begin{equation}
\label{eq: two}
\text { Score Mean}_d=\frac{\sum_{i=d}^{d+1} \text { Score }_i}{N_d}(\mathrm{~d}=0,1,2,3,4,5)
\end{equation}

\begin{align}
\label{eq: three}
\text{Score Std}_d &= \sqrt{\frac{1}{N_d} \sum_{i=d}^{N_d}\left(\text{Score}_i - \text{Score Mean}_d\right)^2} \\
(\mathrm{d}&=0,1,2,3,4,5)
\end{align}

\equref{eq: two} segments the range from 0 to 60 meters into intervals of 10 meters each, designating each as the \(d^{th}\) interval. Within each interval, the sum of the confidence scores of detected objects is divided by the number of detected objects in that interval to calculate the average confidence score. Similarly, \equref{eq: three} computes the standard deviation of confidence scores within each \(d^{th}\) interval, divided by the interval's average confidence score. A quadratic trend line is derived by adjusting the six average confidence scores obtained from \equref{eq: two} within the standard deviations specified in \equref{eq: three}. This process determines the values of the parameters \( \alpha \), \( \beta \), \( \gamma \) and \( k \) for \equref{eq: one}.
\section{Experiments}
\label{sec:experiment}

\subsection{Evaluation Metrics}
Two distinct datasets were utilized for the evaluation: the Kitti 3D Object Detection Dataset \cite{geiger2012we}, a publicly available dataset designed for qualitative assessment of our algorithms, and the custom urban road dataset specifically gathered for quantitative analysis.

The custom urban road dataset comprises data acquired from the Hyundai Elec-city, outfitted with six Velodyne 32CH LiDAR sensors. This vehicle conducted data collection drives around Blue House and Gyeongbokgung in Seoul, capturing diverse environmental conditions ranging from clear to rainy weather scenarios. Additionally, for further data diversity, the Kia Carnival, equipped with four Velodyne 32CH LiDAR sensors, drove in the foggy and rainy environments of Gangneung roads, acquiring a comprehensive dataset encompassing various weather conditions.

\subsection{Quantitative Results}

Our algorithm is a car class that accounts for 93\% of Kiti 3D object detection datasets and quantitatively evaluates whether the 3D object detection model is suitable for autonomous vehicle systems. 3D detector's suitability for autonomous vehicles, ensuring safe and efficient driving from Recall, Precision, and mAP is quantitatively indicated by stable or improving mAP values with balanced Recall and Precision. This implies the model's accurate object detection and recognition efficiency while minimizing false positives, ensuring safe vehicle operation. Therefore, Performance metrics such as Recall, Precision, their Trade-off, and mAP were employed to evaluate the algorithm's impact on 3D detectors before and after our algorithm's implementation.

\begin{table}[t]
\centering
\caption{ Recall, Precision and mAP comparison for PointPillars \cite{lang2019pointpillars} applying various single threshold and adaptive threshold}
\label{tab: one}
\resizebox{0.95\columnwidth}{!}{
\begin{tabular}{c|c|c|c|c|c}
\toprule
\multicolumn{2}{c|}{Method} & Recall & Precision & \textbf{Trade-Off} & mAP   \\ \midrule
\multirow{3}{*}{Single Threshold} & 0.3 & 0.895  & 0.646     & 0.249     & 77.28 \\
 & 0.5       & 0.807  & 0.847     & 0.040     & 77.29 \\
 & 0.7       & 0.655  & 0.943     & 0.288     & 77.49 \\ \midrule
\multicolumn{2}{c|}{\textbf{Ours}}     & 0.786  & 0.813     & \textbf{0.025}     & 77.29 \\ \midrule
\end{tabular}
}
\end{table}

% \usepackage{multirow}
\begin{table}[t]
\centering
\caption{Recall, Precision and mAP comparison for various 3D 
object detection models applying single threshold 0.5 and adaptive thresholding. }
\label{tab: two}
\resizebox{1\columnwidth}{!}{
\begin{tabular}{c|c|c|c|c|c}
\toprule
                                  & 3D Detectors & Recall & Precision & \textbf{Trade-Off}      & mAP   \\ \midrule
\multirow{4}{*}{{Single Threshold}} & PointPillars \cite{lang2019pointpillars} & 0.807  & 0.847      & 0.040                   & 77.28 \\
                                  & SECOND \cite{yan2018second}  & 0.808  & 0.856      & 0.048                   & 78.62 \\
                                  & PointRCNN \cite{shi2019pointrcnn}    & 0.899  & 0.848      & 0.051                   & 78.74 \\
                                  & PV-RCNN \cite{shi2020pv}      & 0.969  & 0.731      & 0.238                   & 79.25 \\ \midrule
\multirow{4}{*}{{\textbf{Ours}}}    & PointPillars \cite{lang2019pointpillars} & 0.786  & 0.813      & \textbf{0.023 (-0.015)} & 77.28 \\
                                  & SECOND \cite{yan2018second}       & 0.792  & 0.823      & \textbf{0.031 (-0.016)} & 78.62 \\
                                  & PointRCNN \cite{shi2019pointrcnn}    & 0.849  & 0.815      & \textbf{0.034 (-0.017)} & 78.73 \\
                                  & PV-RCNN \cite{shi2020pv}      & 0.893  & 0.792      & \textbf{0.101 (-0.137)} & 79.49 \\ \midrule
\end{tabular}
}
\end{table}
\tabref{tab: one} contrasts the outcomes of implementing diverse single thresholds on PointPillars \cite{lang2019pointpillars} with those of the proposed algorithm. Optimal parameter values were established through iterative experimentation. The algorithm's parameters, as indicated in \equref{eq: one}, are \( \alpha \) = -0.00002, \( \beta \) = -0.0061, \( \gamma \) = 0.6828, and \( k\) = 0.6. Our algorithm resulted in a decreased trade-off compared to a single threshold, maintaining comparable mAP performance. 
This highlights the algorithm's ability to fine-tune Recall and Precision for objects near and far, improving Precision for closer objects and increasing Recall for those at extended distances.

Further validation of the algorithm's efficiency was conducted through its implementation to additional LiDAR-based 3D object detection models: SECOND \cite{yan2018second}, PointRCNN \cite{shi2019pointrcnn}, and PV-RCNN \cite{shi2020pv}. These models were subject to performance evaluation based on parameters derived from the PointPillars \cite{lang2019pointpillars}'s experiment. The outcomes with a single 0.5 threshold for each model are tabulated in \tabref{tab: two}, while results leveraging the proposed algorithm are documented in \tabref{tab: two}. As depicted in \tabref{tab: two}, the proposed algorithm for various 3D detectors significantly reduced trade-offs relative to a single threshold, demonstrating that the algorithm improves prediction performances suited for autonomous driving.
 
\subsection{Qualitative Results}
\begin{figure}[t!]
    \centering
    \includegraphics[width=1\columnwidth]{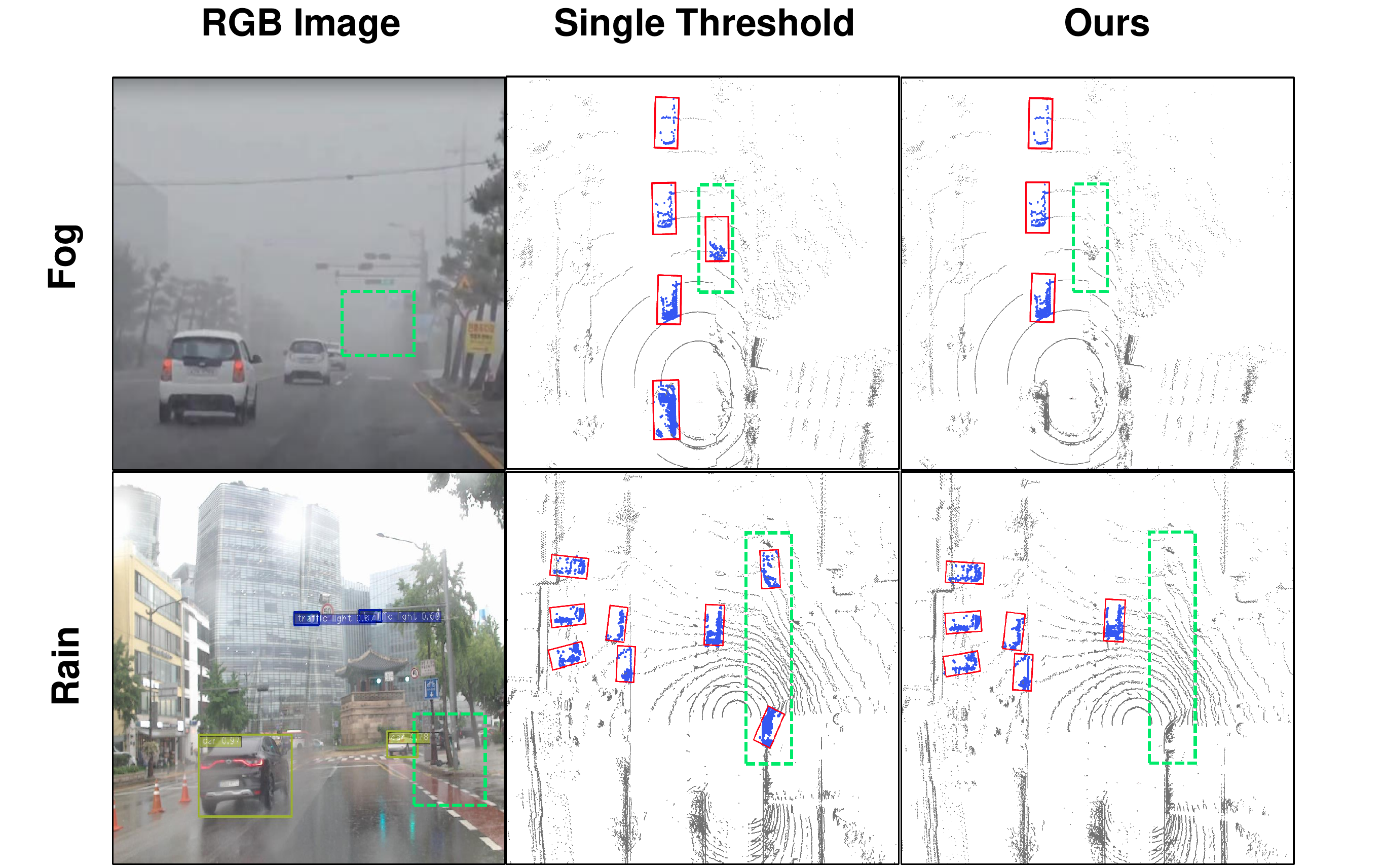}
    \caption{
Results of our algorithm. In challenging urban road scenarios, such as fog and rain, our algorithm enhances the performance of 3D object detection models by reducing false positives and accurately distinguishing vehicles from point clouds caused by adverse weather conditions. This leads to improved overall precision of object detection, ensuring safer driving for autonomous vehicles.} 

\label{fig:Results}
\vspace{-0.5cm}
\end{figure}
Using our urban road dataset, we conducted a qualitative assessment using the PointPillars\cite{lang2019pointpillars} model to address the issue of false positives in various real urban roads and weather conditions. 

\figref{fig: Qualitative Result} shows urban road data acquired in clear weather conditions near Blue House and Gyeongbokgung in Seoul. In the RGB image of \figref{fig: Qualitative Result}., bushes exist horizontally along the right side of the road, from which many point clouds emerge, resembling the shape of vehicles. This leads the model to misidentify bushes as vehicles when applying a single threshold. During actual autonomous driving, such false positives were mistakenly perceived as vehicles changing lanes to the driving lane of the ego-vehicle, causing the vehicle to suddenly stop. When the adaptive thresholding algorithm proposed in this paper was applied to the model, it no longer recognized bushes as vehicles while still accurately detecting vehicles in the left lane and ahead.

\figref{fig:Results}'s fog presents urban road data captured in environments with heavy fog around Gangneung. In the RGB image of \figref{fig:Results}, vehicles can be observed in the left lane, indicating the presence of significant fog. A characteristic of fog data is its appearance at a similar height to LiDAR, resembling smoke-like point noise. Consequently, when a single threshold is applied, false positives of densely fogged areas as floating vehicles in the air can be observed. Our algorithm does not recognize the falsely identified fog, while vehicles existing in the left lane continue to be accurately detected.

\figref{fig:Results}'s rain presents urban road data acquired in heavy rainfall around Blue House and Gyeongbokgung in Seoul. The RGB image in \figref{fig:Results} indicates a situation with substantial precipitation. Although vehicles exist in the left lane, it is observed that no vehicles are present in the driving lane ahead or the right lane. False positives of vehicles occur due to the large amount of point clouds reflected by the heavy rain, and similar to \figref{fig: Qualitative Result}, false positives caused by bushes on the right can be observed. Our algorithm improves Precision at close distances for false positives caused by rain and bushes while vehicles in the left lane remain accurately detected.

%TABLE

%FIGURE
\section{Conclusion \& Discussion}
\label{sec:conclusion}

This research presented an adaptive thresholding algorithm for post-processing in 3D object detection models, which are pivotal for the perception module of autonomous vehicles, a key component of Field Robotics. Through comparative experiments on an open dataset between single thresholds and our algorithm, we've demonstrated that our approach enhances the robustness of the perception module. It effectively minimizes false positives by dynamically adjusting thresholds according to the distance from the autonomous vehicle. 
Subsequent experiments demonstrated the algorithm's effectiveness and adaptability across diverse 3D object detection frameworks. Additionally, qualitative results in the custom dataset have verified the algorithm's practical effectiveness under dynamic urban roads and adverse weather conditions, such as fog and rain. This shows our algorithm's ability to enhance detection accuracy and robustness for the perception module of autonomous driving.

For future work, we will extend the applicability of our methodology beyond LiDAR-based 3D object detection models to those leveraging cameras, radars, and sensor fusion, comprehensive studies on the distinct characteristics of each sensor type. 
Moreover, this algorithm has the potential to evolve into a learning-based approach for real-time adaptation to various scenarios.

% \newpage
% \newpage

%\section*{ACKNOWLEDGMENT}
\balance
\small
\bibliographystyle{IEEEtranN} %citeauthor
\bibliography{string-short,references}

\end{document}